\pdfoutput=1

\documentclass[11pt]{article}

\usepackage{naacl2021}

\usepackage{times}
\usepackage{latexsym}

\usepackage[T1]{fontenc}

\usepackage[utf8]{inputenc}

\usepackage{url}
\usepackage{hyperref}
\usepackage{caption}
\usepackage{graphicx}
\usepackage{booktabs}
\usepackage{subcaption}
\usepackage{mathtools, cuted}
\usepackage{dsfont}
\usepackage{array}
\usepackage{enumitem}
\usepackage{amsmath,amsfonts,amssymb,amsthm, bm}

\DeclareMathOperator*{\argmax}{arg\,max}
 
\usepackage{microtype}

\usepackage{fancyhdr}
\title{gComm: An environment for \\ investigating generalization in Grounded Language Acquisition}

\author{Rishi Hazra \\
  Indian Institute of Science, Bangalore \\
  \texttt{rishihazra@iisc.ac.in} \\\And
  Sonu Dixit \\
  Indian Institute of Science, Bangalore \\
  \texttt{sonudixit@iisc.ac.in} \\}

\begin{document}
\maketitle
\thispagestyle{fancy}

\begin{abstract}
    gComm\footnote{codes \& baselines: \href{https://github.com/SonuDixit/gComm}{https//github.com/SonuDixit/gComm}} is a step towards developing a robust platform to foster research in grounded language acquisition in a more challenging and realistic setting. It comprises a 2-d grid environment with a set of agents (a stationary speaker and a mobile listener connected via a communication channel) exposed to a continuous array of tasks in a partially observable setting. The key to solving these tasks lies in agents developing linguistic abilities and utilizing them for efficiently exploring the environment. The speaker and listener have access to information provided in different modalities, i.e. the speaker's input is a natural language instruction that contains the target and task specifications and the listener's input is its grid-view. Each must rely on the other to complete the assigned task, however, the only way they can achieve the same, is to develop and use some form of communication. gComm provides several tools for studying different forms of communication and assessing their generalization.
\end{abstract}




\section{Environment Description}
\label{appendix:environment}
Recently, datasets embodied in action and perception have been used to train models for various tasks \cite{Vries2018TalkTW,mao2018the}. One such dataset is the grounded SCAN (gSCAN) dataset \cite{ruis2020benchmark} which is used for systematic generalization. We base our environment gComm on the gSCAN dataset which is a grounded version of SCAN benchmark \cite{scan_benchmark}. While both these tasks focus on generalization with the meaning grounded in states of a grid-world, there are however, certain key differences between gComm and gSCAN: \textbf{(i)} Firstly, gSCAN focuses on rule-based generalization for navigation tasks, wherein, an agent learns to map a natural language instruction and its corresponding grid-view to a sequence of action primitives. Contrary to that, we present emergent communication as our main theme, using a pair of bots, a stationary speaker and a mobile listener, that process the language instruction and the grid-view respectively; \textbf{(ii)} Secondly, unlike the supervised framework adopted for learning gSCAN tasks, we use a more realistic RL-framework, wherein, the listener learns by exploring its environment and interacting with it. Our environment is conceptually similar to the BabyAI platform \cite{chevalier-boisvert2018babyai}. However, contrary to BabyAI , which focuses on language \textit{learning}, we intend to project gComm as a general purpose platform for investigating generalization from the perspective of grounded language \textit{acquisition} through emergent communication. A companion paper~\cite{DBLP:journals/corr/abs-2012-05011} introduced a intrinsic reward framework to induce compositionality in the emergent language using the gComm environment. In this paper, we intend to expound on the environment details.


\begin{figure}[t]
	\centering
		\includegraphics[width=0.8\linewidth]{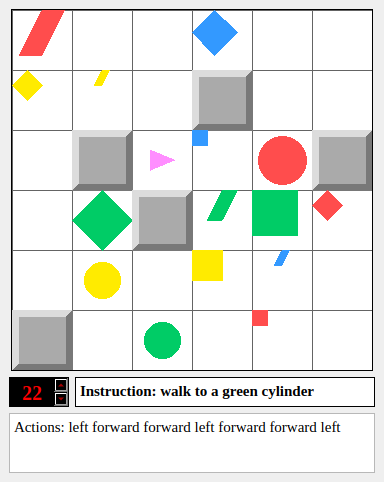}
		\caption{\small gComm Environment}
    \label{figure:comm_gscan}
    \vspace{-0.4cm}
\end{figure}

\paragraph{Object Attributes:} 
The gComm grid-world is populated with objects of different characteristics like shape, color, size and weight. 
\begin{itemize}[leftmargin=*,noitemsep]
    \item \textbf{Shapes:} \textit{circle, square, cylinder, diamond}
    
    \item \textbf{Colors:} \textit{red, blue, yellow, green}
    
    \item \textbf{Sizes:} $1,2,3,4$
    
    \item \textbf{Weights:} \textit{light, heavy}
\end{itemize}

The weight attribute can be fixed corresponding to the object size at the beginning of training. For instance, smaller sized objects are lighter and vice versa. Alternatively, the weight can be set as an independent attribute. In the latter option, the weight is randomly fixed at the start of each episode so that the listener cannot deduce the same from the grid information (object size), and must rely on the speaker.


\subsection{Reinforcement Learning framework}
\label{appendix: rl framework}

\paragraph{Setup:}
In each round, a task is assigned to a stationary Speaker-Bot, the details of which (task and target information) it must share with a mobile Listener-Bot by transmitting a set of messages $m_{i=1}^{n_m}$, via a communication channel. At each time-step $t$, the listener agent selects an action from its action space $\mathcal{A}$, with the help of the received messages $m_{i=1}^{n_m}$ and its local observation (grid-view) $o_t \in \mathcal{O}$. The environment state is updated using the transition function $\mathcal{T}$: $\mathcal{S} \times \mathcal{A} \rightarrow \mathcal{S}$. The environment provides a reward to the agent at each time-step using a reward function $r$: $\mathcal{S} \times \mathcal{A} \rightarrow \mathbb{R}$. The goal of the agent is to find a policy $\bm{\pi}_{\theta}$ : $(\mathcal{O},m_{i=1}^{n_m}) \rightarrow \Delta(\mathcal{A})$ that chooses optimal actions so as to maximize the expected reward, $\mathcal{R} = \mathrm{E}_{\bm{\pi}} [\sum_{t} \gamma^t r^{(t)}]$ where $r^t$ is the reward received by the agent at time-step $t$ and $\gamma \in (0, 1]$ is the discount factor. At the beginning of training, their semantic repertoires are empty, and the speaker and listener must converge on a systematic usage of symbols to complete the assigned tasks thus, giving rise to an original linguistic system.


\paragraph{Observation Space:} 
To encourage communication, gComm provides a partially observable setting in which neither the speaker nor the listener has access to the complete state information. The speaker knows the task and target specifics through the natural language instruction whereas, the listener has access to the grid representation. However, the listener is unaware of either the target object or the task, and therefore must rely on the speaker to accomplish the given task. The observation space of the listener comprises (i) the grid representation; (ii) the messages transmitted by the speaker. 

The natural language instruction is parsed to $\langle\mathrm{VERB}, \{\mathrm{ADJ}_i\}_{i=1}^{3}, \mathrm{NOUN}\rangle$ with the help of an ad hoc semantic parser\footnote{$\mathrm{VERB}$: task; $\mathrm{ADJ}$: object attributes like color, size and weight; $\mathrm{NOUN}$: object shape}. It is then converted to the following 18-d vector representation before being fed to the speaker: \{\textit{1, 2, 3, 4, square, cylinder, circle, diamond, r, b, y, g, light, heavy, walk, push, pull, pickup}\}. Each position represents a bit and is set or unset according to the attributes of the target object and the task. The breakdown of the vector representation is as follows: bits [$0-3$]: target size; bits [$4-7$]: target shape; bits [$8-11$]: target color; bits [$12-13$]: target weight; bits [$14-17$]: task specification.

The grid information can either be a image input of the whole grid or a predefined cell-wise vector representation of the grid. In the latter case, each grid cell in is specified by a 17-d vector representation given by: \{\textit{$1$, $2$, $3$, $4$, square, cylinder, circle, diamond, r, b, y, g, agent, E, S, W, N}\}. The breakdown is as follows: bits [$0-3$]: object size; bits [$4-7$]: object shape; bits [$8-11$]: object color; bit $12$: agent location (is set $=1$ if agent is present in that particular cell, otherwise $0$); bits [$13-16$]: agent direction. For an $obstacle$ or a $wall$, all the bits are set to $1$. 

\paragraph{Action Space:}
The action space comprises eight different actions that the listener agent can perform: \{\textit{left, right, forward, backward, push, pull, pickup, drop}\}. In order to execute the `push', `pull', and `pickup' actions, the agent must navigate to the same cell as that of the object. Upon executing a \textit{pickup} action, the object disappears from the grid. Conversely, an object that has been picked up can reappear in the grid only if a `drop' action is executed in the same episode. Also refer Section~\ref{section: task description} for further details about task descriptions.

\paragraph{Rewards:} 
gComm generates a 0-1 (sparse) reward, i.e., the listener gets a reward of $r = 1$ if it achieves the specified task, otherwise $r = 0$.

\paragraph{Communication:}
Recall that the listener has incomplete information of its state space and is thus unaware of the task and the target object. To address the information asymmetry, the speaker must learn to use the communication channel for sharing information. What makes it more challenging is the fact that the semantics of the transmitted information must be learned in a sparse reward setting, i.e. to solve the tasks, the speaker and the listener must converge upon a common protocol and use it systematically with minimal feedback at the end of each round. 


\subsection{Task Description}
\label{section: task description}

\textbf{(i) Walk} to a target object
\textbf{(ii) Push} a target object in the forward direction.
\textbf{(iii) Pull} a target object in the backward direction.
\textbf{(iv) Pickup} a target object.
\textbf{(v) Drop} the picked up object.

Additionally, there are modifiers associated with verbs, for instance: \textit{pull the red circle twice}. Here, \textit{twice} is a numeral adverb and must be interpreted to mean two consecutive `pull' actions. When an object is picked up, it disappears from the grid and appears only if a `drop' action is executed in the subsequent time-steps. However, no two objects can overlap. It should be noted that while defining tasks, it is ensured that the target object is unique.

\paragraph{Target and Distractor objects:}
Cells in the grid-world are populated with objects divided into two classes: the \textit{target} object and the \textit{distractor} objects. The distractors either have the same color or the same shape (or both) as that of the target. Apart from these, some random objects distinct from the target can also be sampled using a parameter \textit{other\_objects\_sample\_percentage}. The listener and the objects may spawn at any random location on the grid.

\paragraph{Levels:} In addition to the simple grid-world environment comprising target and distractor objects, the task difficulty can be increased by generating obstacles and mazes. The agent is expected to negotiate the complex environment in a sparse reward setting. The number of obstacles and the maze density can be adjusted.

\paragraph{Instruction generation:}
Natural language instructions are programmatically generated based on predefined lexical rules and the specified vocabulary. At the beginning of training, the user specifies the kind of verb (transitive or intransitive), noun (object shape), and adjectives (object weight, size, color). Note, that he instruction templates are fixed, and as such, cannot handle ambiguities in natural language.

\subsection{Communication}
\label{appendix: communication details}
gComm endows the agents with the ability to To encourage communication, gComm provides a partially observable setting in which neither the speaker nor the listener has access to the complete state information. The speaker knows the task and target specifics through the natural language instruction whereas, the listener has access to the grid representation. However, the listener is unaware of either the target object or the task, and hence, it must rely on the speaker to accomplish the given task. The observation space of the listener comprises (i) the grid representation; (ii) the messages transmitted by the speaker. communicate. This forms a crucial step in addressing the partial observability problem and encouraging language acquisition. Above all, gComm provides several tools for an in-depth analysis of grounded communication protocols and their relation to the generalization performance.

\begin{figure}[t]
	\centering
		\includegraphics[width=0.7\linewidth]{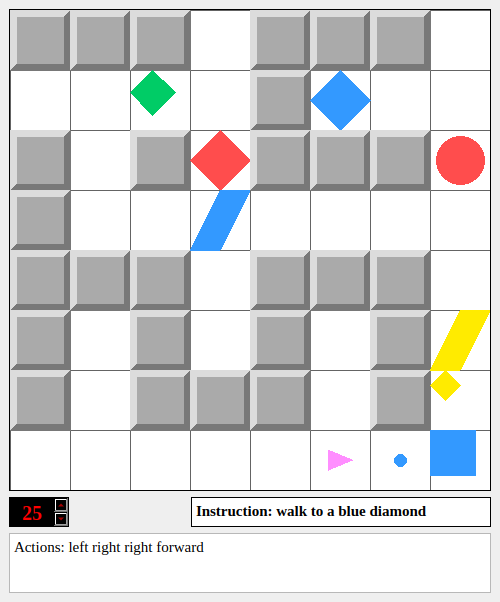}
		\caption{\small Maze-grid. The maze complexity and density are user-defined parameters. The agent is required to negotiate the obstacles while performing the given task.}
    \label{figure:mazegrid}
\end{figure}

\paragraph{Communication Channel:}
\label{appendix:communication_channel}
The communication can be divided into two broad categories.
\begin{itemize}[leftmargin=*,noitemsep]
    \item \textbf{Discrete}: 
    Discrete messages can either be binary (processed using Gumbel-Softmax \cite{JangEtAl:2017:CategoricalReparameterizationWithGumbelSoftmax}) or one-hot (processed using Categorical distribution)\footnote{The use of discrete latent variables render the neural network non-differentiable. The Gumbel Softmax gives a differentiable sample from a discrete distribution by approximating the hard one-hot vector into a soft version. For one-hot vectors, we use Relaxed one-hot Categorical sampling. Since we want the communication to be discrete, we employ the \textit{Straight-Through} trick for both binary and one-hot vectors.}. Discrete messages are associated with a temperature parameter $\tau$. 
    
    \item \textbf{Continuous}: As opposed to discrete messages, continuous signals are real-valued. Theoretically speaking, each dimension in the message can carry 32-bits of information (32-bit floating point). These messages don't pose the same kind of information bottleneck as their discrete counterpart, however, they are not as interpretable.
\end{itemize}

Apart from these, the communication channel can be utilized to compare against the following baseline implementations readily available in the gComm environment. These baselines not only enable us to investigate the efficacy of the emergent communication protocols, but also provides quantitative insights into the learned communication abilities. 

\label{baselines}
\begin{itemize}[leftmargin=*,noitemsep]
    \item \textbf{Random Speaker}: In this baseline, the speaker transmits a set of random symbols to the listener which it must learn to ignore (and focus only on its local observation). 
    
    \item \textbf{Fixed Speaker}: Herein, the speaker's transmissions are masked with a set of \textit{ones}. Intuitively, this baseline provides an idea of whether communication is being used in the context of the given task (whether the speaker actually influences the listener or just appears to do so).
    
    \item \textbf{Perfect Speaker}: This baseline provides an illusion of a perfect speaker by directly transmitting the input concept encoding, hence, acting as an upper bound for comparing the learned protocols.
    
    \item \textbf{Oracle Listener}: For each cell, we zero-pad the grid encoding with an extra bit, and set it ($=1$) for the cell containing the target object. Thus, the listener has complete information about the target in context of the distractors. This baseline can be used as the upper limit of performance.
\end{itemize}

\paragraph{Channel parameters:}
The communication channel is defined using the following parameters:
\begin{itemize}[leftmargin=*,noitemsep]
    \item Message Length: Length of the message vector $d_m$ sets a limit on the vocabulary size, i.e. higher the message length, larger is the vocabulary size. For instance, for discrete (binary) messages, the vocabulary size is given by $|\mathcal{V}| = 2^{d_m}$. Note, that a continuous message can transmit more information compared to a discrete message of the same length.
    
    \item Information Rate or the number of messages $n_m$ transmitted per round of communication.
\end{itemize}
These constitute the channel capacity, $|\mathrm{C}| = \mathrm{c}_{d_m}^{n_m}$.

\paragraph{Setting:}
Communication can either be modelled in form of \textit{cheap talk} or \textit{costly signalling}. In the latter case, each message passing bears a small penalty to encourage more economic and efficient communication protocols. Alternatively, the communication can either be unidirectional (message passing from speaker to listener only) or bidirectional (an interactive setting wherein message passing happens in either direction). gComm uses an unidirectional cheap talk setting.

\begin{figure}[t]
	\centering
		\includegraphics[width=\linewidth]{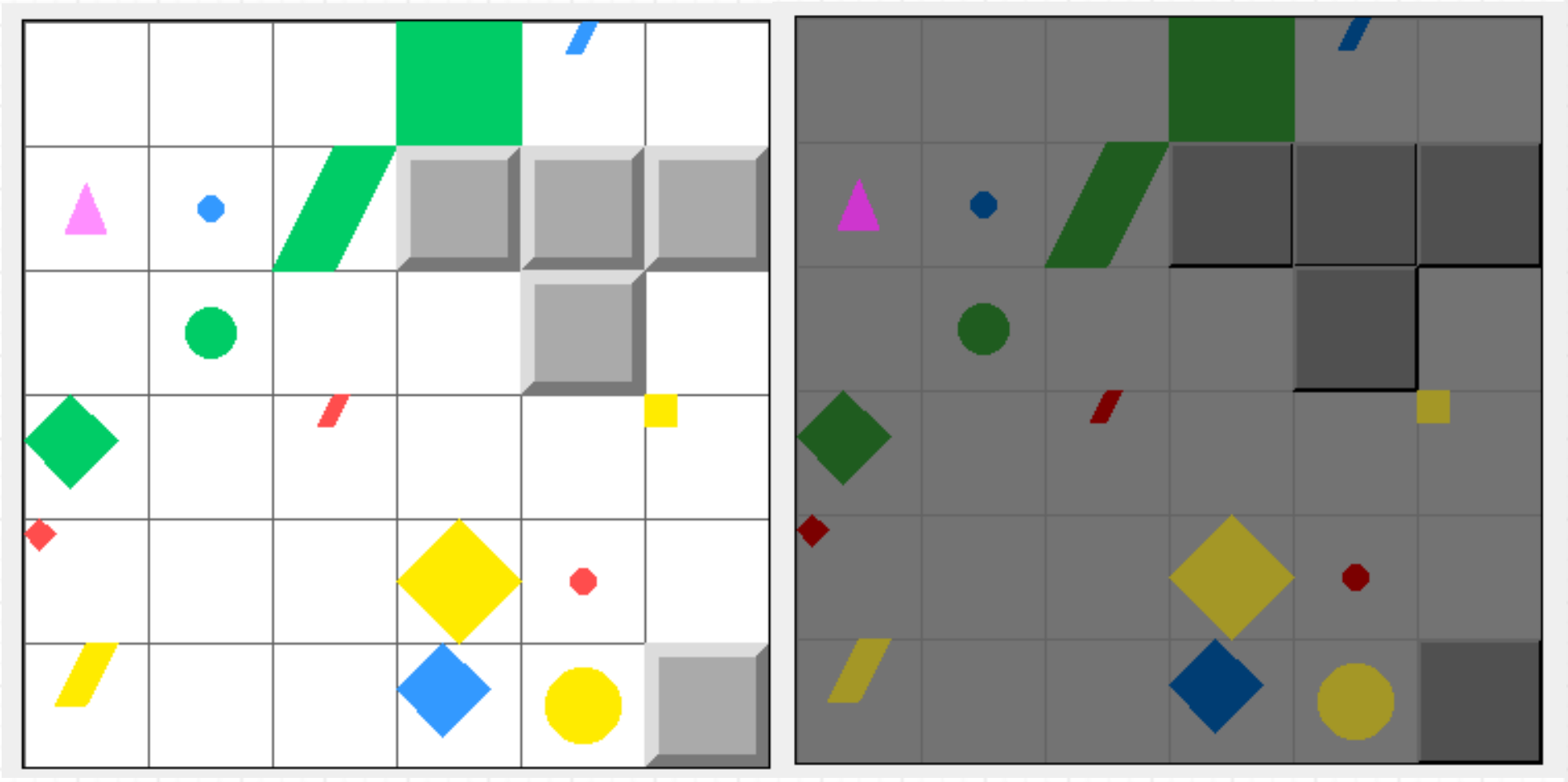}
		\caption{Lights Out}
    \label{figure:lights_off}
\end{figure}

\subsection{Metrics:}
\label{appendix:communication_metrics}
In order to induce meaningful communication protocols, the speaker must transmit useful information, correlated with its input (\textit{positive signalling}). At the same time, the listener must utilize the received information to alter its behavior and hence, its actions (\textit{positive listening}). In alignment with the works of \cite{Lowe2019OnTP}, we incorporate the following metrics in our environment to assess the evolved communication protocols. 

\begin{itemize}[leftmargin=*,noitemsep]
    \item \textbf{Positive signalling}: 
    Context independence (CI) is used as an indicator of positive signalling. It captures the statistical alignment between the input concepts and the messages transmitted by the speaker and is given by:
    \begin{align*}
        \forall c \in \mathcal{C}: m_c = \argmax_m p_{cm}(c|m) \\
        CI(p_{mc}, p_{cm}) = \frac{1}{|\mathcal{C}|} \sum_c p_{cm}(c|m_c)p_{mc}(m_c|c)
    \end{align*}
   
   Both $p_{cm}(c|m)$ and $p_{mc}(m|c)$ are calculated using a translation model by saving ($m,c$) pairs and running it in both directions. Since each concept element $c$ should be mapped to exactly one message $m$, CI will be high when the $p_{cm}(c|m)$ and $p_{mc}(m|c)$ are high.\\
    
    \item \textbf{Positive listening}: We use Causal Influence of Communication (CIC) of the speaker on the listener as a measure of positive listening. It is defined as the mutual information between the speaker's message and the listener's action $I(m,a_t)$. Higher the CIC, more is the speaker's influence on the listener's actions, thus, indicating that the listener is utilizing the messages.\\
    
    \item \textbf{Compositionality}: Compositionality is measured using the topographic similarity (topsim) metric \cite{10.1162/106454606776073323}. Given two pairwise distance measures, i.e. one in the concept (input) space $\Delta_{\mathcal{C}}^{ij}$ and another in the message space $\Delta_{\mathcal{M}}^{ij}$, topsim is defined as the correlation coefficient calculated between $\Delta_{\mathcal{C}}^{ij}$ and $\Delta_{\mathcal{M}}^{ij}$. Higher topsim indicates more compositionality.
\end{itemize}

\subsection{Additional features}
\label{section: additional features}
We introduce a \textit{lights out} feature in the gComm environment through which the grid (including all its objects) is subjected to varying illuminations (Figure~\ref{figure:lights_off}). The feature can be activated randomly in each episode and presents a challenging situation for the agent where it is required to navigate the grid using its memory of the past observation. Note that this feature is useful only when used with an image input as the grid representation.


\begin{table}[t]
	\small
	 \begin{center}
		 \begin{tabular}{ >{\centering\arraybackslash}m{1.8cm} >{\centering\arraybackslash}m{2.2cm} >{\centering\arraybackslash}m{1.7cm}}
			 \toprule
			 \textbf{Task} & \textbf{Baseline} & \textbf{Convergence Rewards}\\[0.4ex] 
			 \midrule
			 \textbf{Walk} & \begin{tabular}{>{\centering\arraybackslash}m{2.2cm}>{\centering\arraybackslash}m{1.5cm}>{\centering\arraybackslash}m{1.5cm}>{\centering\arraybackslash}m{1.5cm}>{\centering\arraybackslash}m{1.5cm}} Simple Speaker & $0.70$ \\ \midrule Random Speaker & $0.40$ \\ \midrule Fixed Speaker & $0.43$ \\ \midrule Perfect Speaker & $0.95$ \\ \midrule Oracle Listener & $0.99$ \end{tabular}\\
			 \midrule
			 \textbf{PUSH} \& \textbf{PULL} & \begin{tabular}{>{\centering\arraybackslash}m{2.2cm}>{\centering\arraybackslash}m{1.5cm}>{\centering\arraybackslash}m{1.5cm}>{\centering\arraybackslash}m{1.5cm}>{\centering\arraybackslash}m{1.5cm}} Simple Speaker & $0.55$ \\ \midrule Random Speaker & $0.19$ \\ \midrule Fixed Speaker & $0.15$ \\ \midrule Perfect Speaker & $0.85$ \\ \midrule Oracle Listener & $0.90$ \end{tabular}\\
			 \bottomrule
		 \end{tabular}
	 \end{center}
	 \caption{\small Comparison of baseline convergence rewards [\textbf{Task: Walk}, params: \{comm\_type: categorical, num\_episodes: 200000, episode\_len: 10, num\_msgs: 3, msg\_len: 4, num\_actions: 4 (left, right, forward, backward), type\_grammar: simple\_intrans, weights: light, enable\_maze: False, grid\_size: $4\times4$, distractors: 4, grid\_input\_type: vector\}][ \textbf{Task: Push/Pull}, params: \{comm\_type: categorical, num\_episodes: 400000, episode\_len: 10, num\_msgs: 3, msg\_len: 4, num\_actions: 6 (left, right, forward, backward, push, pull), type\_grammar: simple\_trans, weights: light, enable\_maze: False, grid\_size: $4\times4$, distractors: 2, grid\_input\_type: vector\}]. Note, that these rewards were recorded over a set of $100$ validation episodes.}
	 \label{tab_results1}
 \end{table}
 
 
\section{Related Work}

\paragraph{Emergent Communication:} With regard to emergent communication, so far, most existing works are limited to analyzing simple referential games \cite{Lewis1969-LEWCAP-4} in simulated environments, where a speaker communicates the input (object's shape and color) to a stationary listener which, then, tries to classify the reconstructed messages from a list of classes  \cite{kottur-etal-2017-natural,HavrylovEtAl:2017:EmergenceOfLanguageWithMultiAgentGamesLearningToCommunicateWithSequencesOfSymbols,CaoEtAl:2018:EmergentCommunicationThroughNegotiation,NEURIPS2019_b0cf188d}. These games do not involve world state manipulation and generally comprise elementary inputs with limited attributes, thus, restricting the scope of language usage. gComm introduces an additional challenge for the listener to navigate and manipulate objects to achieve the transmitted goal.

\paragraph{Visual Navigation:} The problem of navigating in an environment based on visual perception, by mapping the visual input to actions, has
long been studied in vision and robotics. The tasks are either specified implicitly via rewards~\cite{8100252}, or are explicitly conditioned on the goal state (Goal-conditioned Reinforcement Learning)~\cite{zhu2017icra,10.5555/3327546.3327593,NEURIPS2019_c8cc6e90}. In contrast, gComm tasks are specified using natural language and involves unidirectional messages from a \textit{task-aware} speaker to a \textit{state-aware} listener. 

\paragraph{Embodied Learning:} Recent works on embodied learning include (but are not limited to) using embodied agents to complete tasks specified by natural language in a simple mazeworld \cite{10.5555/3305890.3305956}, Embodied Question Answering~\cite{8575449} and Embodied Language Grounding by mapping 2D scenes to 3D~\cite{Prabhudesai_2020_CVPR}. We also intend to project gComm as a embodied communication environment where the listener agent is required to ground the messages to its corresponding visual input and associate them with actions (\textit{push a red circle twice} suggests that the red circle is heavy and the listener needs to perform two consecutive ``push" actions to move it.)

\paragraph{Instruction Execution:} These approaches focus on natural language understanding to map instructions to actions \cite{branavan-etal-2009-reinforcement,10.5555/2900423.2900560,10.5555/2900423.2900661}. However, in gComm, the listener agent doesn't have direct access to the natural language instruction hence, it focuses on mapping transmitted messages from the speaker to actions. The challenge is to address the information bottleneck, i.e., given a limited channel capacity, the speaker must learn to convey the required task and target specifics to the listener based on the input instruction.

\paragraph{Visual Question Answering:} In VQA, agents are required to answer natural language questions based on a fixed view of the environment (image or video) \cite{7410636,7780870,fukui-etal-2016-multimodal,8954214}. However, unlike gComm, the agents cannot actively perceive or manipulate objects.

\section{Discussion}
\label{section:discussion}
We compared a Simple Speaker (speaker transmitting one-hot messages) with the baselines given in Section~\ref{baselines} for \textbf{(i) Walk} task wherein the listener is required to walk to a target object; \textbf{(ii)} \textbf{Push} $+$ \textbf{Pull} task wherein the listener is required to push or pull a target object. The grid we used was of size $4 \times 4$ with no obstacles. Moreover, we used 5 objects (4 distractors $+$ 1 target) for (i) and 3 objects (2 distractors $+$ 1 target) for (ii). The number of messages were set at 3 (i.e., one messages each for task, shape and color).

We present our analysis based on the results from Table~\ref{tab_results1}.
\begin{itemize}[leftmargin=*,noitemsep]
    \item Simple Speaker outperforms the Fixed and Random baselines.
    \item Perfect Speaker performs as well as Oracle Listener.
    \item Oracle Listener had the fastest convergence ($\approx \frac{1}{5}$ of the episodes taken by Simple Speaker), followed by Perfect Speaker ($\approx \frac{1}{2}$ of the episodes taken by Simple Speaker).
    \item Fixed Speaker baseline converges faster than the Random Speaker baseline which suggests that the Listener learns to ignore messages if they remain fixed over time.
\end{itemize}

\bibliography{biblio}

\begin{thebibliography}{28}
\expandafter\ifx\csname natexlab\endcsname\relax\def\natexlab#1{#1}\fi

\bibitem[{Antol et~al.(2015)Antol, Agrawal, Lu, Mitchell, Batra, Zitnick, and
  Parikh}]{7410636}
Stanislaw Antol, Aishwarya Agrawal, Jiasen Lu, Margaret Mitchell, Dhruv Batra,
  C.~Lawrence Zitnick, and Devi Parikh. 2015.
\newblock \href {https://doi.org/10.1109/ICCV.2015.279} {Vqa: Visual question
  answering}.
\newblock In \emph{2015 IEEE International Conference on Computer Vision
  (ICCV)}, pages 2425--2433.

\bibitem[{Branavan et~al.(2009)Branavan, Chen, Zettlemoyer, and
  Barzilay}]{branavan-etal-2009-reinforcement}
S.R.K. Branavan, Harr Chen, Luke Zettlemoyer, and Regina Barzilay. 2009.
\newblock \href {https://www.aclweb.org/anthology/P09-1010} {Reinforcement
  learning for mapping instructions to actions}.
\newblock In \emph{Proceedings of the Joint Conference of the 47th Annual
  Meeting of the {ACL} and the 4th International Joint Conference on Natural
  Language Processing of the {AFNLP}}, pages 82--90. Association for
  Computational Linguistics.

\bibitem[{Brighton and Kirby(2006)}]{10.1162/106454606776073323}
Henry Brighton and Simon Kirby. 2006.
\newblock \href {https://doi.org/10.1162/106454606776073323} {Understanding
  linguistic evolution by visualizing the emergence of topographic mappings}.
\newblock \emph{Artif. Life}.

\bibitem[{Cao et~al.(2018)Cao, Lazaridou, Lanctot, Leibo, Tuyls, and
  Clark}]{CaoEtAl:2018:EmergentCommunicationThroughNegotiation}
Kris Cao, Angeliki Lazaridou, Marc Lanctot, Joel~Z Leibo, Karl Tuyls, and
  Stephen Clark. 2018.
\newblock \href {https://openreview.net/forum?id=Hk6WhagRW} {Emergent
  communication through negotiation}.
\newblock In \emph{International Conference on Learning Representations (ICLR)
  2018}.

\bibitem[{Chen and Mooney(2011)}]{10.5555/2900423.2900560}
David~L. Chen and Raymond~J. Mooney. 2011.
\newblock \href {https://dl.acm.org/doi/10.5555/2900423.2900560} {Learning to
  interpret natural language navigation instructions from observations}.
\newblock In \emph{Proceedings of the Twenty-Fifth AAAI Conference on
  Artificial Intelligence}, page 859–865.

\bibitem[{Chevalier-Boisvert et~al.(2019)Chevalier-Boisvert, Bahdanau, Lahlou,
  Willems, Saharia, Nguyen, and Bengio}]{chevalier-boisvert2018babyai}
Maxime Chevalier-Boisvert, Dzmitry Bahdanau, Salem Lahlou, Lucas Willems,
  Chitwan Saharia, Thien~Huu Nguyen, and Yoshua Bengio. 2019.
\newblock \href {https://openreview.net/forum?id=rJeXCo0cYX} {Baby{AI}: First
  steps towards grounded language learning with a human in the loop}.
\newblock In \emph{International Conference on Learning Representations (ICLR)
  2019}.

\bibitem[{Das et~al.(2018)Das, Datta, Gkioxari, Lee, Parikh, and
  Batra}]{8575449}
Abhishek Das, Samyak Datta, Georgia Gkioxari, Stefan Lee, Devi Parikh, and
  Dhruv Batra. 2018.
\newblock \href {https://doi.org/10.1109/CVPRW.2018.00279} {Embodied question
  answering}.
\newblock In \emph{2018 IEEE/CVF Conference on Computer Vision and Pattern
  Recognition Workshops (CVPRW)}, pages 2135--213509.

\bibitem[{Fukui et~al.(2016)Fukui, Park, Yang, Rohrbach, Darrell, and
  Rohrbach}]{fukui-etal-2016-multimodal}
Akira Fukui, Dong~Huk Park, Daylen Yang, Anna Rohrbach, Trevor Darrell, and
  Marcus Rohrbach. 2016.
\newblock \href {https://doi.org/10.18653/v1/D16-1044} {Multimodal compact
  bilinear pooling for visual question answering and visual grounding}.
\newblock In \emph{Proceedings of the 2016 Conference on Empirical Methods in
  Natural Language Processing}, pages 457--468. Association for Computational
  Linguistics.

\bibitem[{Gupta et~al.(2017)Gupta, Davidson, Levine, Sukthankar, and
  Malik}]{8100252}
Saurabh Gupta, James Davidson, Sergey Levine, Rahul Sukthankar, and Jitendra
  Malik. 2017.
\newblock \href {https://doi.org/10.1109/CVPR.2017.769} {Cognitive mapping and
  planning for visual navigation}.
\newblock In \emph{2017 IEEE Conference on Computer Vision and Pattern
  Recognition (CVPR)}, pages 7272--7281.

\bibitem[{Havrylov and
  Titov(2017)}]{HavrylovEtAl:2017:EmergenceOfLanguageWithMultiAgentGamesLearningToCommunicateWithSequencesOfSymbols}
Serhii Havrylov and Ivan Titov. 2017.
\newblock \href {https://dl.acm.org/doi/10.5555/3294771.3294976} {Emergence of
  language with multi-agent games: Learning to communicate with sequences of
  symbols}.
\newblock In \emph{Proceedings of the 31st International Conference on Neural
  Information Processing Systems}, NIPS'17, page 2146–2156.

\bibitem[{Hazra et~al.(2020)Hazra, Dixit, and
  Sen}]{DBLP:journals/corr/abs-2012-05011}
Rishi Hazra, Sonu Dixit, and Sayambhu Sen. 2020.
\newblock \href {https://arxiv.org/abs/2012.05011} {Infinite use of finite
  means: Zero-shot generalization using compositional emergent protocols}.
\newblock \emph{CoRR}, abs/2012.05011.

\bibitem[{Jang et~al.(2017)Jang, Gu, and
  Poole}]{JangEtAl:2017:CategoricalReparameterizationWithGumbelSoftmax}
Eric Jang, Shixiang Gu, and Ben Poole. 2017.
\newblock \href {https://openreview.net/forum?id=rkE3y85ee} {Categorical
  reparameterization with gumbel-softmax}.
\newblock In \emph{International Conference on Learning Representations (ICLR)
  2017}.

\bibitem[{Kottur et~al.(2017)Kottur, Moura, Lee, and
  Batra}]{kottur-etal-2017-natural}
Satwik Kottur, Jos{\'e} Moura, Stefan Lee, and Dhruv Batra. 2017.
\newblock \href {https://www.aclweb.org/anthology/D17-1321} {Natural language
  does not emerge {`}naturally{'} in multi-agent dialog}.
\newblock In \emph{Proceedings of the 2017 Conference on Empirical Methods in
  Natural Language Processing}, pages 2962--2967. Association for Computational
  Linguistics.

\bibitem[{Lake and Baroni(2018)}]{scan_benchmark}
Brenden Lake and Marco Baroni. 2018.
\newblock \href {http://proceedings.mlr.press/v80/lake18a.html} {Generalization
  without systematicity: On the compositional skills of sequence-to-sequence
  recurrent networks}.
\newblock In \emph{Proceedings of the 35th International Conference on Machine
  Learning}, Proceedings of Machine Learning Research. PMLR.

\bibitem[{Lewis(1969)}]{Lewis1969-LEWCAP-4}
David~K. Lewis. 1969.
\newblock \emph{Convention: A Philosophical Study}.
\newblock Wiley-Blackwell.

\bibitem[{Li and Bowling(2019)}]{NEURIPS2019_b0cf188d}
Fushan Li and Michael Bowling. 2019.
\newblock \href
  {https://proceedings.neurips.cc/paper/2019/file/b0cf188d74589db9b23d5d277238a929-Paper.pdf}
  {Ease-of-teaching and language structure from emergent communication}.
\newblock In \emph{Advances in Neural Information Processing Systems},
  volume~32.

\bibitem[{Lowe et~al.(2019)Lowe, Foerster, Boureau, Pineau, and
  Dauphin}]{Lowe2019OnTP}
Ryan Lowe, Jakob Foerster, Y-Lan Boureau, Joelle Pineau, and Yann Dauphin.
  2019.
\newblock \href {https://dl.acm.org/doi/10.5555/3306127.3331757} {On the
  pitfalls of measuring emergent communication}.
\newblock In \emph{Proceedings of the 18th International Conference on
  Autonomous Agents and MultiAgent Systems}, AAMAS '19, page 693–701.
  International Foundation for Autonomous Agents and Multiagent Systems.

\bibitem[{Mao et~al.(2019)Mao, Gan, Kohli, Tenenbaum, and Wu}]{mao2018the}
Jiayuan Mao, Chuang Gan, Pushmeet Kohli, Joshua~B. Tenenbaum, and Jiajun Wu.
  2019.
\newblock \href {https://openreview.net/forum?id=rJgMlhRctm} {The
  neuro-symbolic concept learner: Interpreting scenes, words, and sentences
  from natural supervision}.
\newblock In \emph{International Conference on Learning Representations (ICLR)
  2019}.

\bibitem[{Nair et~al.(2018)Nair, Pong, Dalal, Bahl, Lin, and
  Levine}]{10.5555/3327546.3327593}
Ashvin Nair, Vitchyr Pong, Murtaza Dalal, Shikhar Bahl, Steven Lin, and Sergey
  Levine. 2018.
\newblock \href {https://dl.acm.org/doi/10.5555/3327546.3327593} {Visual
  reinforcement learning with imagined goals}.
\newblock In \emph{Proceedings of the 32nd International Conference on Neural
  Information Processing Systems}, NIPS'18, page 9209–9220.

\bibitem[{Nasiriany et~al.(2019)Nasiriany, Pong, Lin, and
  Levine}]{NEURIPS2019_c8cc6e90}
Soroush Nasiriany, Vitchyr Pong, Steven Lin, and Sergey Levine. 2019.
\newblock \href
  {https://proceedings.neurips.cc/paper/2019/file/c8cc6e90ccbff44c9cee23611711cdc4-Paper.pdf}
  {Planning with goal-conditioned policies}.
\newblock In \emph{Advances in Neural Information Processing Systems},
  volume~32. Curran Associates, Inc.

\bibitem[{Oh et~al.(2017)Oh, Singh, Lee, and Kohli}]{10.5555/3305890.3305956}
Junhyuk Oh, Satinder Singh, Honglak Lee, and Pushmeet Kohli. 2017.
\newblock \href {https://dl.acm.org/doi/10.5555/3305890.3305956} {Zero-shot
  task generalization with multi-task deep reinforcement learning}.
\newblock In \emph{Proceedings of the 34th International Conference on Machine
  Learning - Volume 70}, ICML'17, page 2661–2670. JMLR.org.

\bibitem[{Prabhudesai et~al.(2020)Prabhudesai, Tung, Javed, Sieb, Harley, and
  Fragkiadaki}]{Prabhudesai_2020_CVPR}
Mihir Prabhudesai, Hsiao-Yu~Fish Tung, Syed~Ashar Javed, Maximilian Sieb,
  Adam~W. Harley, and Katerina Fragkiadaki. 2020.
\newblock \href
  {https://openaccess.thecvf.com/content_CVPR_2020/html/Prabhudesai_Embodied_Language_Grounding_With_3D_Visual_Feature_Representations_CVPR_2020_paper.html}
  {Embodied language grounding with 3d visual feature representations}.
\newblock In \emph{Proceedings of the IEEE/CVF Conference on Computer Vision
  and Pattern Recognition (CVPR)}.

\bibitem[{Ruis et~al.(2020)Ruis, Andreas, Baroni, Bouchacourt, and
  Lake}]{ruis2020benchmark}
Laura Ruis, Jacob Andreas, Marco Baroni, Diane Bouchacourt, and Brenden~M Lake.
  2020.
\newblock \href
  {https://proceedings.neurips.cc/paper/2020/file/e5a90182cc81e12ab5e72d66e0b46fe3-Paper.pdf}
  {A benchmark for systematic generalization in grounded language
  understanding}.
\newblock In \emph{Advances in Neural Information Processing Systems},
  volume~33, pages 19861--19872. Curran Associates, Inc.

\bibitem[{Shah et~al.(2019)Shah, Chen, Rohrbach, and Parikh}]{8954214}
Meet Shah, Xinlei Chen, Marcus Rohrbach, and Devi Parikh. 2019.
\newblock \href {https://doi.org/10.1109/CVPR.2019.00681} {Cycle-consistency
  for robust visual question answering}.
\newblock In \emph{2019 IEEE/CVF Conference on Computer Vision and Pattern
  Recognition (CVPR)}, pages 6642--6651.

\bibitem[{Tapaswi et~al.(2016)Tapaswi, Zhu, Stiefelhagen, Torralba, Urtasun,
  and Fidler}]{7780870}
Makarand Tapaswi, Yukun Zhu, Rainer Stiefelhagen, Antonio Torralba, Raquel
  Urtasun, and Sanja Fidler. 2016.
\newblock \href {https://doi.org/10.1109/CVPR.2016.501} {Movieqa: Understanding
  stories in movies through question-answering}.
\newblock In \emph{2016 IEEE Conference on Computer Vision and Pattern
  Recognition (CVPR)}, pages 4631--4640.

\bibitem[{Tellex et~al.(2011)Tellex, Kollar, Dickerson, Walter, Banerjee,
  Teller, and Roy}]{10.5555/2900423.2900661}
Stefanie Tellex, Thomas Kollar, Steven Dickerson, Matthew~R. Walter,
  Ashis~Gopal Banerjee, Seth Teller, and Nicholas Roy. 2011.
\newblock \href {https://dl.acm.org/doi/10.5555/2900423.2900661} {Understanding
  natural language commands for robotic navigation and mobile manipulation}.
\newblock In \emph{Proceedings of the Twenty-Fifth AAAI Conference on
  Artificial Intelligence}, page 1507–1514.

\bibitem[{Vries et~al.(2018)Vries, Shuster, Batra, Parikh, Weston, and
  Kiela}]{Vries2018TalkTW}
H.~D. Vries, Kurt Shuster, Dhruv Batra, D.~Parikh, J.~Weston, and Douwe Kiela.
  2018.
\newblock \href {https://arxiv.org/abs/1807.03367} {Talk the walk: Navigating
  new york city through grounded dialogue}.
\newblock \emph{ArXiv}.

\bibitem[{Zhu et~al.(2017)Zhu, Mottaghi, Kolve, Lim, Gupta, Fei-Fei, and
  Farhadi}]{zhu2017icra}
Yuke Zhu, Roozbeh Mottaghi, Eric Kolve, Joseph~J. Lim, Abhinav Gupta,
  Li~Fei-Fei, and Ali Farhadi. 2017.
\newblock \href {https://ieeexplore.ieee.org/document/7989381} {{Target-driven
  Visual Navigation in Indoor Scenes using Deep Reinforcement Learning}}.
\newblock In \emph{{IEEE International Conference on Robotics and Automation}}.

\end{thebibliography}
\bibliographystyle{acl_natbib}
\end{document}